\documentclass[conference]{IEEEtran}
\usepackage[english]{babel}
\usepackage[T1]{fontenc}
\usepackage{epsfig}
\IEEEoverridecommandlockouts
\usepackage{tabu}
\usepackage{float}
\usepackage{amsmath}
\usepackage{amssymb}
\usepackage{amsthm}
\usepackage[keeplastbox]{flushend}
\usepackage{cite}
\usepackage{subfigure}
\usepackage{epstopdf}
\usepackage{multirow}
\usepackage{caption}
\usepackage{mathtools}
\usepackage{xfrac}
\usepackage{units}
\usepackage[keeplastbox]{flushend}
\usepackage[hyphens]{url}
\usepackage{array}
\newcommand{\beql}[1]{\begin{equation}\label{#1}}
\newcommand{\eeq}{\end{equation}}

\newcommand{\be}{\begin{equation}}
\newcommand{\ee}{\end{equation}}
\newcommand{\ba}{\begin{array}}
\newcommand{\ea}{\end{array}}

\begin{document}
%\begindocument

%\title{Achievable Rates in Coexisting Cognitive Radio Networks}
\title{Efficient Privacy Preserving Edge Computing Framework for Image Classification}

\author{\IEEEauthorblockN{Omobayode Fagbohungbe, Sheikh Rufsan Reza, Xishuang Dong, Lijun Qian}
\IEEEauthorblockA{Center of Excellence in Research and Education for Big Military Data Intelligence (CREDIT) \\
%%Department of Electrical and Computer Engineering\\
 Prairie View A\&M University, Texas A\&M University System \\
 Prairie View, TX 77446, USA\\
Email: ofagbohungbe, sreza@student.pvamu.edu, xidong@pvamu.edu, liqian@pvamu.edu}
\thanks{© 2021 IEEE.  Personal use of this material is permitted.  Permission from IEEE must be obtained for all other uses, in any current or future media, including reprinting/republishing this material for advertising or promotional purposes, creating new collective works, for resale or redistribution to servers or lists, or reuse of any copyrighted component of this work in other works.}
}

\maketitle

\begin{abstract}
In order to extract knowledge from the large data collected by edge devices, traditional cloud based approach that requires data upload may not be feasible due to communication bandwidth limitation as well as privacy and security concerns of end users. To address these challenges, a novel privacy preserving edge computing framework is proposed in this paper for image classification. Specifically, autoencoder will be trained unsupervised at each edge device individually, then the obtained latent vectors will be transmitted to the edge server for the training of a classifier. This framework would reduce the communications overhead and protect the data of the end users. Comparing to federated learning, the training of the classifier in the proposed framework does not subject to the constraints of the edge devices, and the autoencoder can be trained independently at each edge device without any server involvement. Furthermore,  the privacy of the end users' data is protected by transmitting latent vectors without additional cost of encryption. Experimental results provide insights on the image classification performance vs. various design parameters such as the data compression ratio of the autoencoder and the model complexity.
\end{abstract}


\section{Introduction and Motivation}
\label{sec:introduction}
\begin{figure}[htbp]
	\centering
	\includegraphics[width=9cm]{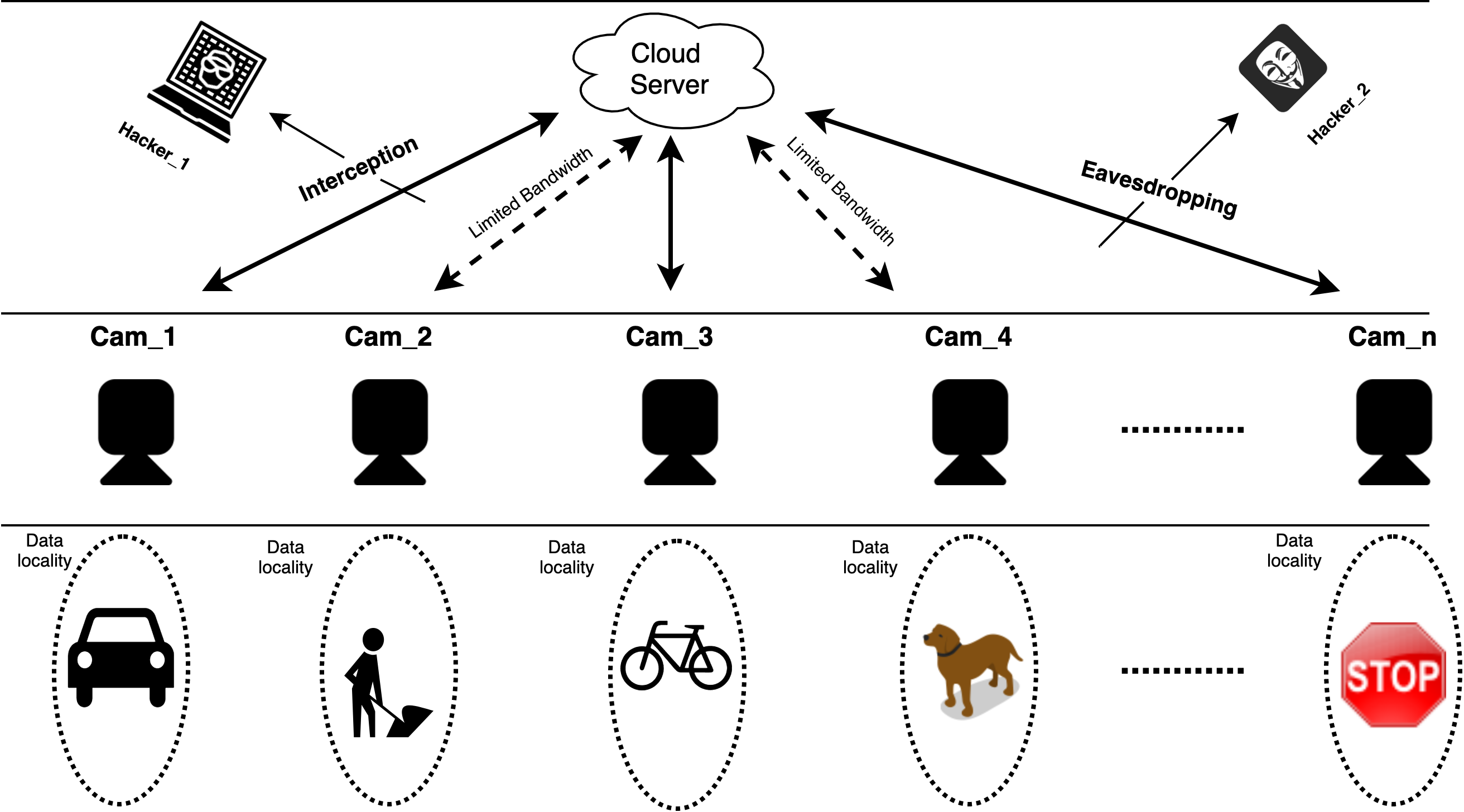}
	\caption{Challenges incurred when uploading all data from edge devices to the cloud.}
	\label{fig:ChallengeUploadAllData}
\end{figure}

Emerging Technologies such as the Internet of Things (IoT) and 5G networks will add a huge number of devices and new services, as a result, a huge amount of data will be generated in real time. One of the important data types is image data, since
many applications involve images and videos such as in video surveillance.
In order to take advantage of the ``big image data'', data analytics must be performed to extract knowledge from the data. 
One way to handle the data would be uploading all data from edge devices to the cloud or remote data centers for processing and knowledge extraction~\cite{aws}. However, as highlighted in Figure~\ref{fig:ChallengeUploadAllData}, there are several factors that may render this practice infeasible: 1) The sheer volume of the images may overwhelm the uplink with limited bandwidth;  2) The uplink may not be always available especially when wireless communications is used due to weather (e.g., for mmWave), distance, or jamming; 3) Proprietary images may need encryption that introduce additional delay; 4) The end users may have concerns about the security and privacy of their images, thus they may not agree to upload raw images that may contain private information. Furthermore, uploading is subject to eavesdropping, interceptions, or other unauthorized access.
%%% figure for motivation
%%% motivationAIOT

\begin{figure}[htbp]
	 \centering
    	 \includegraphics[width=9cm]{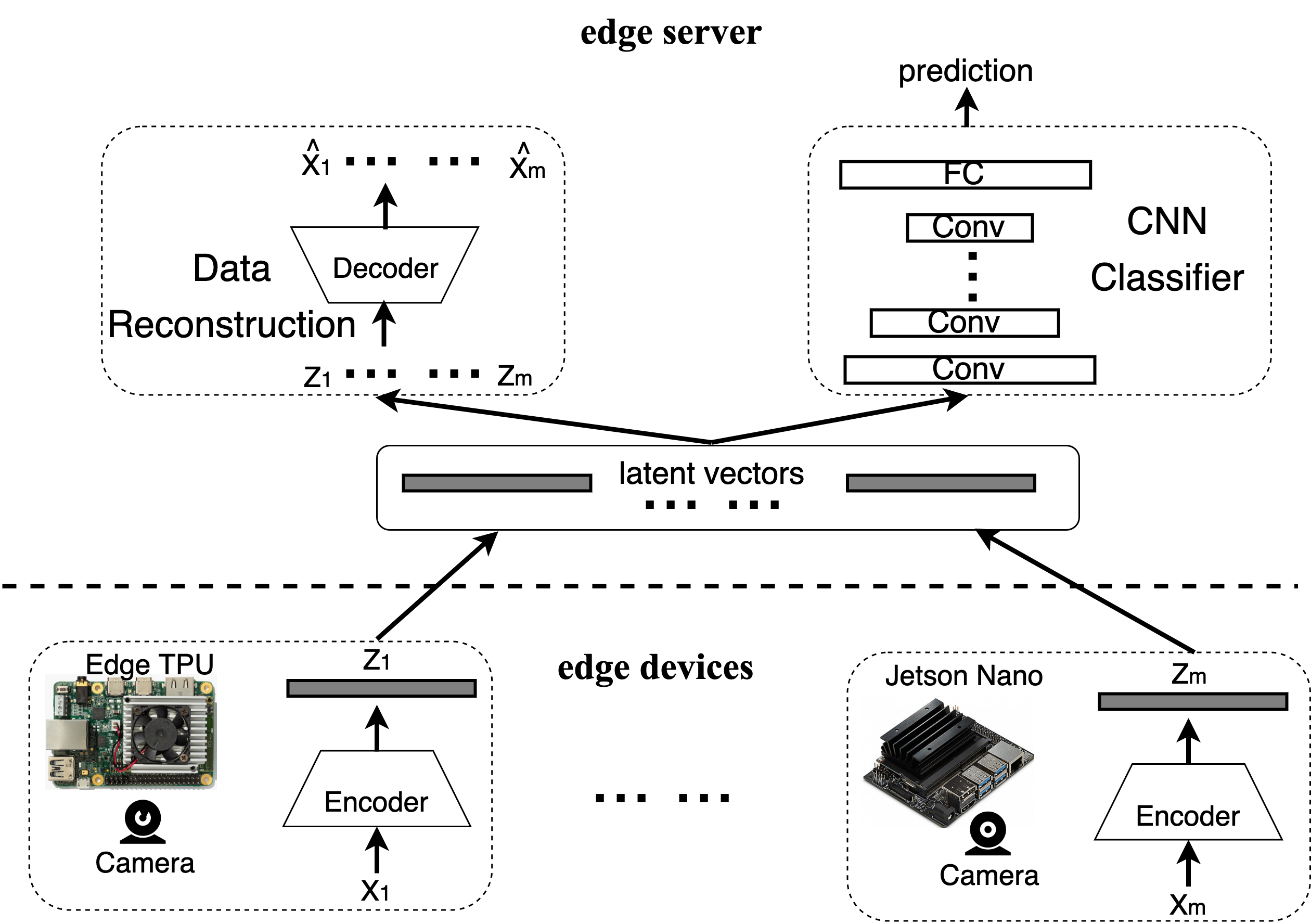}
      	\caption{The proposed efficient privacy preserving framework for image classification in edge computing systems. Here $x_i$ is the raw image, $z_i$ is the compressed latent vector, and $\hat{x}_i$ is the reconstructed image.}
    \label{fig:framework}
\end{figure}

In order to address these challenges, a novel efficient privacy preserving framework for image classification in edge computing systems is proposed in this paper. Specifically, the large raw data will be processed locally (at the edge) by pre-trained autoencoder; then instead of uploading the raw image, only compressed latent vector that contain critical features learned from the raw image will be uploaded through the access point or hub to the cloud for further processing. The proposed framework is highlighted in Figure~\ref{fig:framework}. It is demonstrated in our experiments that the learning performance of extracting knowledge at the cloud has very little degradation when the compression ratio is not large (e.g., below 16 in our test cases). Furthermore, the raw images can be reconstructed with very small error at the cloud using the pre-trained decoder if needed. 

Comparing to traditional source coding (e.g., zip), using autoencoder has the following advantages: 1) Instead of only reducing the redundancy in the raw data as in source coding or traditional data compression, autoencoder will extract critical features in the raw data and encode the features in a compact form (the latent vector), in other words, the encoder performs initial learning at the edge devices; 2) 
In addition to compressing the data, autoencoder also ``encrypt'' the data by transforming the raw data into latent vectors, which enhance the security of data. For example, a zipped file can easily be unzipped by an adversary if not encrypted; on the contrary, an adversary could not reconstruct the raw data from the latent vector without knowing the structure (e.g., number of layers, number of nodes in each layer) and all the weights of the pre-trained autoencoder (the decoder part to be exact), in fact, it is shown in~\cite{AutoencoderNeurology2018} that autoencoder provides a similar level of security to normal encryption - assuming that the decoder is not shared; (3) Even if the edge device is captured by an adversary, it is very difficult for the adversary to deduce the decoder part from the encoder part on the edge device.

%%% comparison to federated learning
The proposed framework has some similar characteristics such as taking advantage of large diverse data from many edge devices and data locality at each device as in federated learning~\cite{McMahanMRA16, KonecnyFederated, MultiObjFederatedLearning, FederatedMachineLearningYangQiang2019}.
However, comparing to federated learning, the proposed framework has the following advantages:
\begin{enumerate}
\item In federated learning, the server and the end users (edge devices) train the same model. As a result, the complexity of the model is constrained by the computing capability and storage of the edge device. On the contrary, \emph{in the proposed framework, the training of the classifier is done at the cloud server only, thus it can be very deep and complex if needed, and  it is not subject to the constraints of the edge devices}.
\item In federated learning, the edge device must rely on the server to update the gradients and train the model. \emph{In the proposed framework, the training of the autoencoder can be done independently at each edge device without any server involvement}.
\item In federated learning, the privacy of the end users' data is protected by applying differential privacy schemes~\cite{DifferentiallyPrivateFederatedLearning} or through secure aggregation~\cite{SecureAggregationforFederatedLearning}, thus introduce additional cost due to encryption or secret sharing. \emph{In the proposed framework, the privacy of the end users' data is protected by transmitting latent vectors without additional cost of encryption}.
\end{enumerate}

The proposed framework is explained in Section~\ref{sec:framework}. Section~\ref{sec:experiment} gives the design and implementation details. Experimental results and analysis are given in Section~\ref{sec:results}. Further discussions and related works are reviewed in Section~\ref{sec:discussion}.  Section~\ref{sec:conclusion} concludes the paper.

\section{Proposed Framework}
\label{sec:framework}
The proposed efficient privacy preserving framework for image classification in edge computing systems is shown in Figure~\ref{fig:framework}. It has two levels: the edge devices and the edge server. It is assumed that the nodes of the edge devices contain sensors such as cameras and embedded computing devices such as Google edge TPU~\cite{TPU} or NVIDIA Jetson Nano~\cite{jetson}. The edge server is assumed to have strong computational capacity and large storage.

The edge segment of the framework mainly contains the various edge devices of interest and the pre-trained encoder. The server mainly contains the hub, the pre-trained classifier and the pre-trained decoder. We only consider supervised learning in this paper and it is assumed that the training dataset is labeled.

The data from each of the edge devices are passed to the corresponding encoder attached to it. Unique pre-trained encoder is used for each of the edge devices in order to take advantage of the data locality at each device. The function of the pre-trained encoder, which is in the inference mode, is to extract the most important and critical features in the data. The encoder also ensures dimension reduction of the input data by a pre-determined amount. The extracted critical features (latent vectors, or feature maps when the data are images) are then transmitted to the hub at the server.  The two major functions at the server are the classification task and the data reconstruction task (recover a copy the original image from the latent vectors). In other words,  the latent vectors are  input to the pre-trained classifier for prediction and they are also input to the corresponding decoder at the server for the reconstruction of the images. 

\begin{figure}[htbp]
	 \centering
    	 \includegraphics[width=5.5cm]{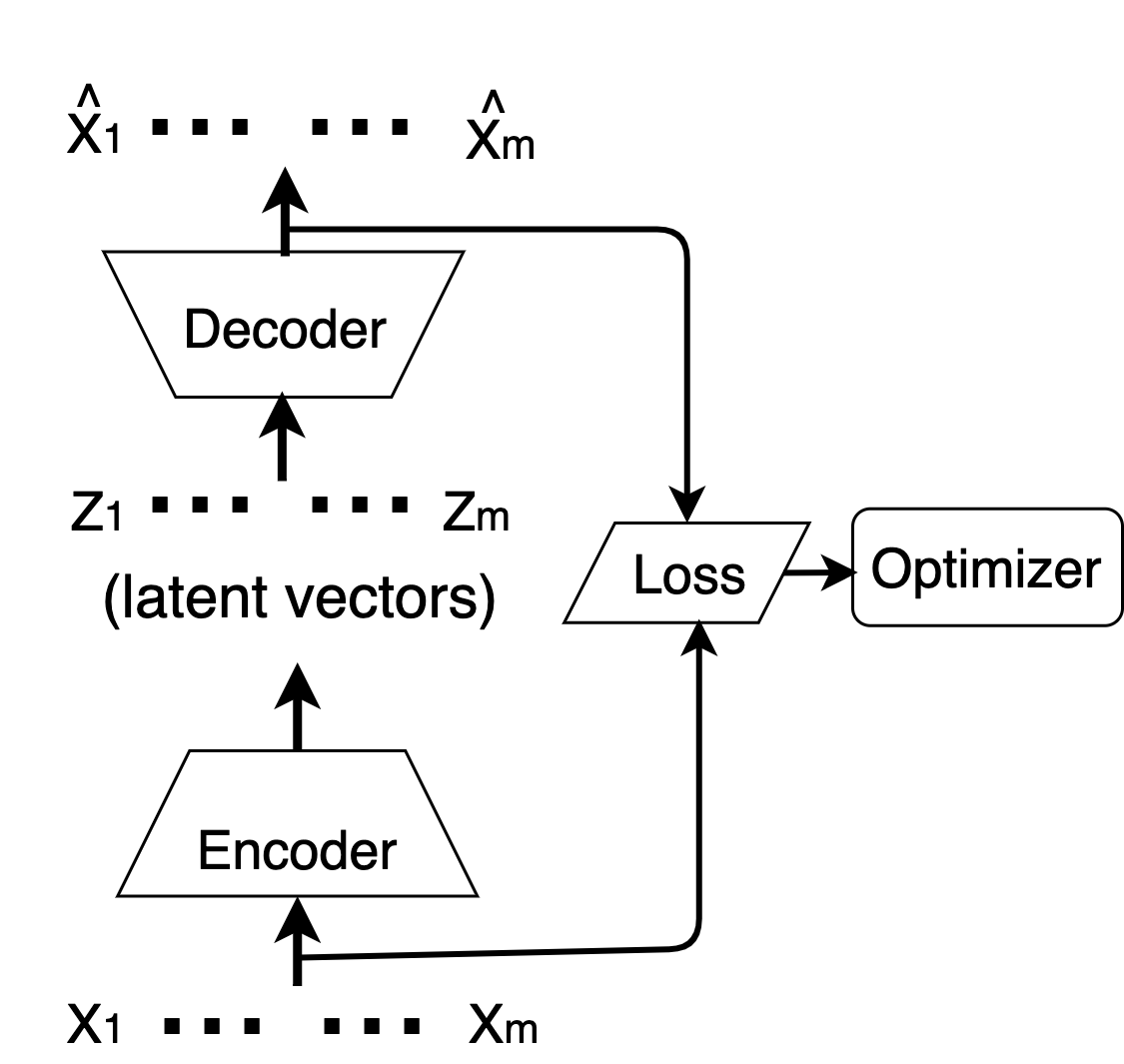}
      	\caption{The training for the proposed autoencoder at edge device.}
    \label{fig:AutoencoderTraining}
\end{figure}

\begin{figure}[htbp]
	 \centering
    	 \includegraphics[width=6.5cm]{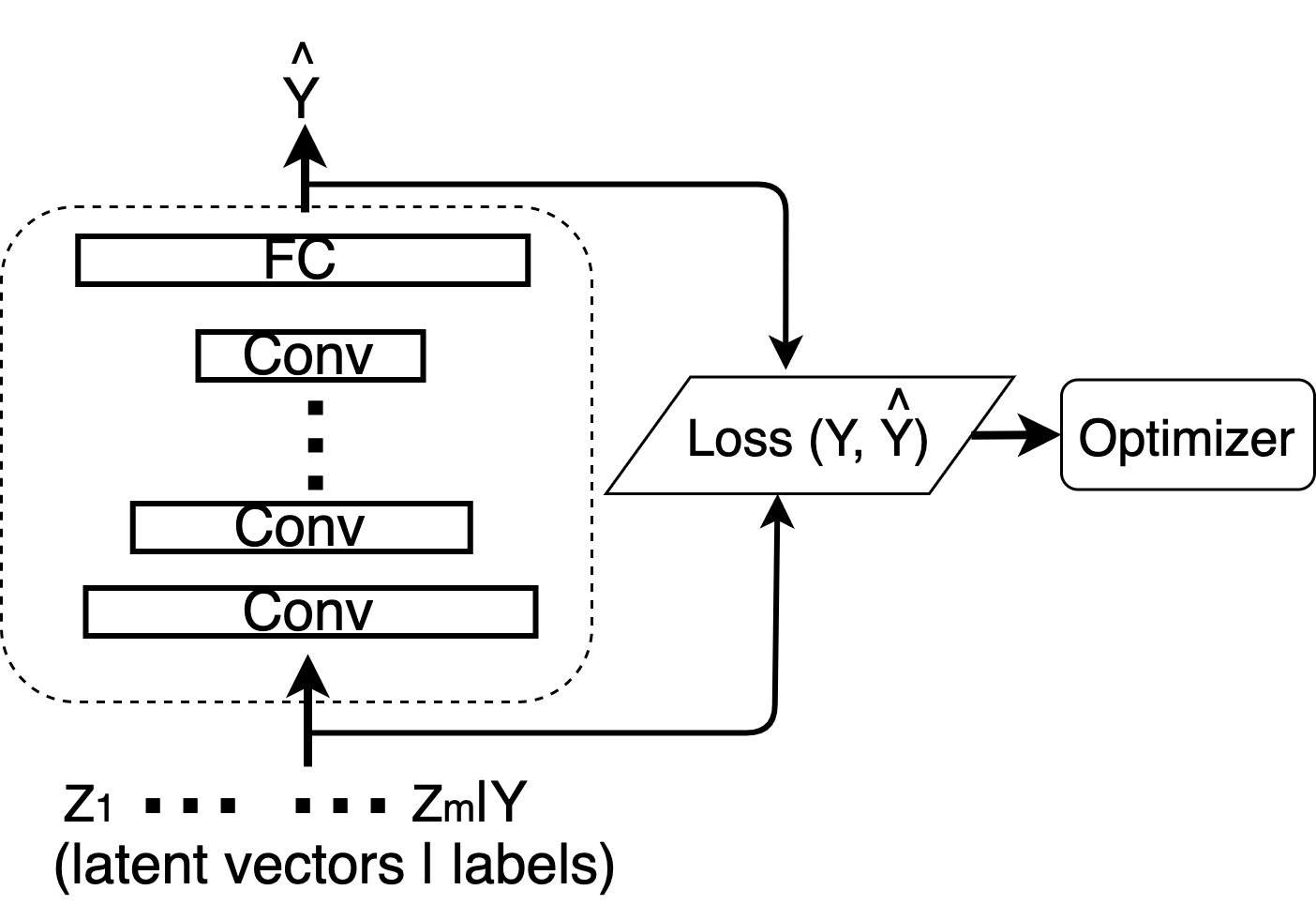}
      	\caption{The training for the proposed CNN classifier at the server.}
    \label{fig:CNNtraining}
\end{figure}

%%% add a Table here for the three datasets
\begin{table*}
\centering
\caption{Information on the CIFAR10 and ImageNet (IMGNET\-A and IMGNET\-B) Datasets}
\begin{tabular}{|c|c|c|c|c|c|}
\hline
\textbf{Dataset} & \textbf{Image size} & \textbf{\# of images} & \textbf{Training\:Testing ratio} & \textbf{\# of classes} & \textbf{Comments} \\\hline\hline
{CIFAR10}  & {32*32*3} & 60 000 & 5\:1 & 10 &  \\ %\cline{1-7}
\hline
{IMGNET-A}  & 256*256*3 & 13,000 & 7:3 & 10 & {very different images} \\ %\cline{1-7}
\hline
{IMGNET-B}  & 256*256*3 & 13,000 & 7:3 & 10 & {very similar images} \\ %\cline{1-7}
\hline
\end{tabular}
\label{table:datasetInfo}
\end{table*}

\subsection{Training Stage}
\label{sec:dataset}
The dataset collected at each edge device is used to train an autoencoder for the corresponding device. This is done to take advantage of the data locality at each devices. Autoencoders are generative models where an artificial neural network is trained to reconstruct its own input in an unsupervised way. 
Figure \ref{fig:AutoencoderTraining} illustrates all the components of an autoencoder and the training process. It is made up by two main blocks which are the encoder and the decoder~\cite{metrology, Goodfellow2015DeepL}. The encoder compresses the input $X$ into a low dimensional representation of pre-determined size, called the latent vector denoted by $Z$ that contains the most important features in the data. When the input data are images, $Z$ will be the corresponding feature maps. The decoder then tries to reconstruct the original data from the latent vector $Z$. The reconstructed data obtained at the decoder output is denoted by  $\hat{X}$.  It should be noted that an autoencoder is a lossy network as the original image will not be fully recovered. However, it is expected that the critical features will remain in the recovered image.

The Autoencoder achieves the proper training of the encoder and decoder by minimizing the differences between original input and the reconstructed input.  This is achieved by the use of the mean square error loss function or any other loss function. After the training of the autoencoder, the encoder part of the autoencoder is then extracted, deployed in the inference mode on the edge device, and then used to generate the latent vector $Z$. Hence, the dataset is transformed from $ [X,Y] $ to $[Z,Y]$ where $Y$ are the labels. 

The latent vectors and the corresponding labels are then aggregated at the hub and they are used for training a classifier on the cloud in a supervised manner as shown in Figure~\ref{fig:CNNtraining}. The type of classifier on the cloud is determined by the nature or type of supervised task to be done. The most common type of classifier used for image dataset is the convolutional neural network (CNN) and it uses the cross entropy loss function.

\subsection{Inference Stage}
\label{sec:dataset}
tfttftftf In this stage,the pre-trained encoder, decoder and classifier are deployed in the inference mode. The data $X$ from the edge device is fed to the corresponding pre-trained encoder attached to it. The encoder then transforms the data $X$ to a latent vector $Z$ which represents the most critical feature in $X$. The latent vector $Z$, which is smaller than $X$ by a pre-determined ratio, is then transmitted to the cloud. At the cloud server, the latent vector $Z$ is then fed into the pre-trained classifier and predict a label $\hat{Y}$. In situations where the original data is needed on the cloud, say we want to see a copy of the original image, the latent vector $Z$ is also fed into the input of the corresponding decoder and the estimated data is obtained. 

\section{Experiments}
\label{sec:experiment}
\subsection{Dataset Description}
The result in this work was generated using three different datasets summarized in Table~\ref{table:datasetInfo}. These datasets are from the Canadian Institute For Advanced Research dataset (CIFAR10)~\cite{cifar10} and the ILSVRC (ImageNet) 2012 datasets~\cite{Imagenet}.

\subsubsection{\textbf{Canadian Institute For Advanced Research (CIFAR10)}}
\label{sec:dataset}
This dataset containing 60,000 color images, is a subset of about 80 million labeled but tiny images. The dataset is further divided into 50,000 training samples and 10,000 testing samples. It has about 10 classes which are mutually exclusive and there's no semantic overlaps between images from different classes.

\subsubsection{\textbf{ILSVRC (ImageNet) 2012}}
\label{sec:dataset}
The original ILSVRC 2012 dataset contain about 1.2 million color images of different sizes across about 1,000 classes. The 1,000 classes are either internal or leaf nodes but do not overlap. Two subsets of the ILSVRC 2012 dataset termed IMGNET-A and IMGNET-B were used in this work. Each subset contains about 13,000 images each resized to a dimension $256\times256$, spanning 10 classes. The images in each subset is further divided into training samples and testing samples with ratio 7:3.  The difference between the two subsets lies in the type of nodes they contain. The IMGNET-A subset contains images from 10 different leaf nodes (diverse images) while IMGNET-B contains 10 child nodes from a single leaf node (similar images). 

\subsection{Deep Learning Model Design and Training Strategy}
\label{sec:dataset}
The autoencoder for the edge devices and the classifier at the edge server were chosen  such that  the autoencoder is optimized for feature extraction and the classifier is optimized for image classification. 

\subsubsection{\textbf{Autoencoder Design and Training Strategy}}
\label{sec:dataset}
%The autoencoder contains two part: the encoder and the decoder. The input of an autoencoder is passed through an encoder which extracts  the most important/critical features of the input, reducing the dimension of the input by a pre-determined factor in the process. The output of the encoder serves as the input of the decoder which now attempts to reconstruct the original input from the feature maps. 
% The reconstructed input is never exactly the same as the input due to the lossy nature of the network though their dimensions are the same. By minimizing the difference between the input and the reconstructed input, an autoencoder acts as a powerful feature extractor.
 \begin{figure}[htbp]
	 \centering
    	 \includegraphics[width=5cm]{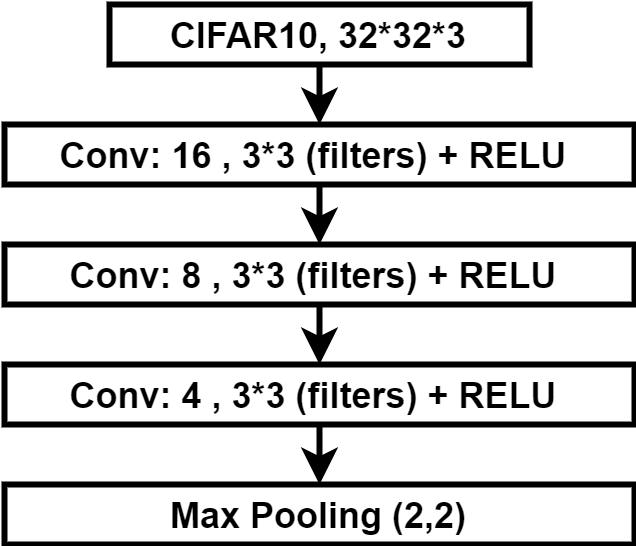}
     	\caption{Details of an encoder model for compression size of 4 using CIFAR10 dataset}
    \label{fig:encoderdetails}
\end{figure}
 %%% add a Table here for Model-A, Model-B and Model-C  
%%% add a Table here for the three datasets
\begin{table*}
\centering
\caption{The deep learning models and the dataset used in training the models}
\begin{tabular}{|c|c|c|c|c|}
\hline
\textbf{} & \textbf{} & \textbf{CIFAR10} & \textbf{IMGNET-A} & \textbf{IMGNET-B}  \\\hline\hline
\multirow{2}{*}{Vanilla Model}  & {Model-A} & x & - & - \\ \cline{2-5}
%\hline
 & Model-B & -& x  & x \\ \cline{2-5}
\hline
{Transfer Model}  & Model-C & - & x & x \\ \cline{2-5}
\hline
\end{tabular}
\label{table:modeldetails}
\end{table*}

\begin{table} [ht] \scriptsize
\caption{The architecture of the vanilla model for CIFAR10 dataset (Model-A) }
\centering
\begin{tabular}{|c|}
\hline
% & & &\multicolumn{6}{c|}{\textbf{Accuracy}} \\ \cline{4-9}
 \textbf{Vanilla Model For Imagenet Dataset} \\  \hline\hline
\multirow{1}{*}{Conv2D, Filter Size = 3*3, No of Filters = 32, Stride = 1*1,No Padding}  \\ \hline
\multirow{1}{*}{Activation Layer (Relu)}  \\ \hline
\multirow{1}{*}{Max Pooling, Pool Size = 2*2, Stride = 1,1,No Padding}   \\ \hline
\multirow{1}{*}{Conv2D, Filter Size = 3*3, No of Filters = 32, No Padding}   \\ \hline
\multirow{1}{*}{Activation Layer (Relu)}  \\ \hline
\multirow{1}{*}{Max Pooling, Pool Size = 2*2,Stride = 1*1,No Padding				
}  \\ \hline
\multirow{1}{*}{Conv2D, Filter Size = 3*3, No of Filters = 32,Stride = 1*1, No Padding				
}  \\ \hline
\multirow{1}{*}{Activation Layer (Relu)}  \\ \hline
\multirow{1}{*}{Max Pooling,Pool Size = 2*2,Stride = 1*1,No Padding				
}  \\ \hline\
\multirow{1}{*}{Flatten}  \\ \hline\
\multirow{1}{*}{Dense(64)				}  \\ \hline
\multirow{1}{*}{Activation( Relu)}  \\ \hline
\multirow{1}{*}{Dropout(0.5)				}  \\ \hline
\multirow{1}{*}{Dense(10)				}  \\ \hline
\multirow{1}{*}{Activation(Softmax)				}  \\ \hline
\end{tabular}
\label{table:cifar10details}
\end{table}

The autoencoder architecture is affected by the type of images and the compression ratio. For instance, the model architecture for CIFAR10 dataset for compression ratios 4 and 8 are different. This also applies for compression ratio 4 for IMGNET-A and CIFAR10 dataset. Hence, different models were developed across several edge devices, compression ratio and datasets. 

 Figure \ref{fig:encoderdetails} shows the model architecture for an autoencoder designed for CIFAR10 dataset for compression ratio of 4. The model contains a mix of convolutional (same padding), max pooling, and upsampling layers. The relu function was used as the activation function for all layers except the last layer where the sigmoid function was used.

 The models were trained from scratch using glorot-uniform method as initializer, mean square error as the cost function and rmsprop optimization algorithm as the optimizer. After the convergence of the autoencoder model during  training process, encoder part of the autoencoder was then used in the inference mode to compress all the images to obtain the latent variables needed to train the classifier.

\subsection{Training Stage}
\label{sec:dataset}
\subsubsection{\textbf{Classifier Design}}
\label{sec:dataset}
The convolutional neural network (CNN) models were used  in this work.
CNNs are well suited for image processing applications and other grid-like data~\cite{Goodfellow2015DeepL}. They are more computationally efficient than the dense deep neural network thus reducing the memory usage. Using the filters, CNNs find and extract meaningful features from the images and preserve spatial relations. Three different CNN classifiers, denoted Model-A, Model-B and Model-C as listed in Table~\ref{table:modeldetails}, were used in this work.

\paragraph{Model-A and Model-B} Model-A and Model-B are considered to be vanilla models because they were trained from scratch. Model-A and Model-B were specifically designed for the original input image and feature maps of CIFAR10 dataset and ImageNet dataset, respectively. The detailed CNN architecture of Model-A and Model-B are shown in Tables \ref{table:cifar10details} and \ref{table:ImagenetModel}, respectively. The models contain a mix of convolutional, max pooling, and fully connected layers and  relu and softmax activation functions.
\begin{table} [ht] \scriptsize
\caption{The architecture of the vanilla model for Imagenet dataset (Model-B)}
\centering
\begin{tabular}{|c|}
\hline
% & & &\multicolumn{6}{c|}{\textbf{Accuracy}} \\ \cline{4-9}
 \textbf{Vanilla Model For CIFAR10 Dataset} \\  \hline\hline
\multirow{1}{*}{Conv2D, Filter Size=3*3, No of Filters=32, Stride=1*1,Padding				
}  \\ \hline
\multirow{1}{*}{Activation Layer (Relu)}  \\ \hline
\multirow{1}{*}{Conv2D, Filter Size=3*3, No of Filters=32, Stride=1*1, Padding	}   \\ \hline
\multirow{1}{*}{Activation Layer (Relu)}  \\ \hline
\multirow{1}{*}{Max Pooling, Pool Size = 2*2,Stride = 1*1, Padding}  \\ \hline
\multirow{1}{*}{Dropout(0.25)}   \\ \hline

\multirow{1}{*}{Conv2D, Filter Size = 3*3, No of Filters = 64,Stride = 1*1, Padding				
}  \\ \hline
\multirow{1}{*}{Activation Layer (Relu)}  \\ \hline
\multirow{1}{*}{Conv2D, Filter Size = 3*3, No of Filters = 64,Stride = 1*1, Padding				
}  \\ \hline
\multirow{1}{*}{Activation Layer (Relu)}  \\ \hline

\multirow{1}{*}{Max Pooling,Pool Size = 2*2,Stride = 1*1, Padding			
}  \\ \hline\
\multirow{1}{*}{Dropout(0.25)				}  \\ \hline
\multirow{1}{*}{Flatten}  \\ \hline\
\multirow{1}{*}{Dense(512)				}  \\ \hline
\multirow{1}{*}{Activation( Relu)}  \\ \hline
\multirow{1}{*}{Dropout(0.5)				}  \\ \hline
\multirow{1}{*}{Dense(10)				}  \\ \hline
\multirow{1}{*}{Activation(Softmax)				}  \\ \hline
\end{tabular}
\label{table:ImagenetModel}
\end{table}
Furthermore, the models also contain some dropout layer in order to prevent over-fitting. The differences between Model-A and Model-B lies in the number of the different layers used and the use of padding in the convolutional layer of Model-A.

\begin{figure}[htbp]
	 \centering
    	 \includegraphics[width=8cm]{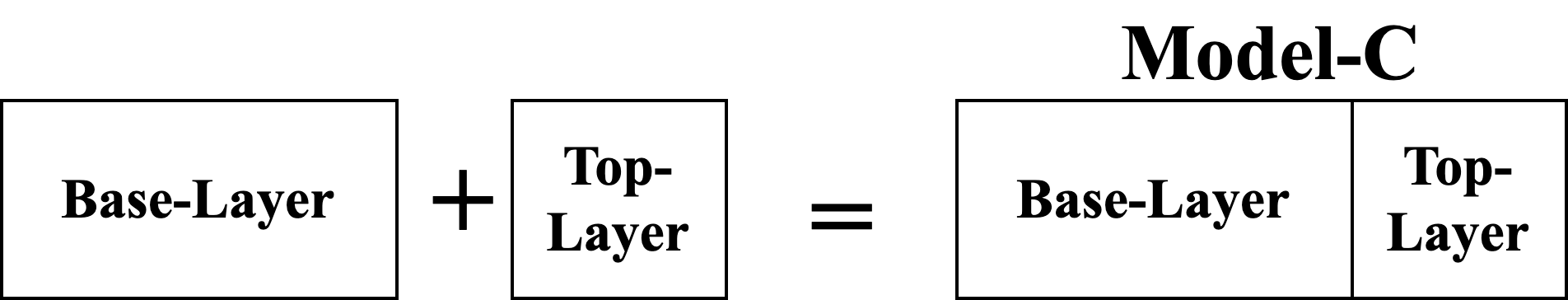}
      	\caption{The Transfer Learning Model Block (Model-C) }
    \label{fig:TF1ARCH}
\end{figure}

\begin{table*}
\centering
\caption{The architecture of the transfer learning model for Imagenet datasets (Model-C)}
\begin{tabular}{|c|c|c|c|c|c|}
\hline
\textbf{} & \textbf{Model\_1} & \textbf{Model\_2} &  \textbf{Model\_3} & \textbf{Model\_4} & \textbf{Model\_5} \\\hline\hline
{Base layer}  & VGG16 & VGG19 & InceptionV3 & InceptionResnetV2 & Resnet50  \\ \cline{2-6}
\hline
\multirow{5}{*}{Top layer}  & Dense(256) & Dense(256) & Dense(256) & Dense(256) & Dense(50)  \\ \cline{2-6}

 	 	    & Activation(Relu) & Activation(Relu) & Activation(Relu) & Activation(Relu) & Activation(Relu) \\ \cline{2-6}
             & Dense(10) & Dense(10) & Dense(10) & Dense(10) & Dense(10) \\ \cline{2-6} 
            & Activation(Softmax) & Activation(Softmax) & Activation(Softmax) & Activation(Softmax) & Activation(Softmax)    \\ \cline{1-6}
           
\end{tabular}
\label{table:modelC}
\end{table*}

The models were trained from scratch with the aim of minimizing the difference between the labels (ground truth) and the predicted labels. This was achieved by the use of glorot-uniform method as initializer, categorical cross entropy as the loss function and adams optimization algorithm as the optimizer. Data augmentation was also used during the training process to mitigate overfitting due to the small quantity of the datasets. It should be stated that each classifier were trained with their respective original image and the feature maps (compressed images).

\paragraph{Model-C} Model-C is a transfer learning based model designed specifically for the ImageNet dataset in this work. The CIFAR10 version was not presented in this work because of poor performance when used with the feature maps (compressed image) of CIFAR10 which can be attributed to its small dimension.
 
The block diagram of the model is shown in Figure \ref{fig:TF1ARCH}. Model-C can be divided into two parts: The base layer and the top layer. The base layer is a pre-trained layer of  other standard deep learning model (without the fully connected layer) that has been trained with data similar to the Imagenet data and achieved better performance. By using this pre-trained models, the excellent feature extracting property of the standard model is being leveraged to achieve better performance. Furthermore, it also complements data augmentation in training decent model in situations where datasets are limited. VGG16, VGG19, InceptionV3, InceptionResnetV2 and Resnet50 pre-trained models~\cite{keras} were used as base models for Model-C.
%% add a ref for the pre-trained models
 
The details of the top layer used for this work is shown in Table~\ref{table:modelC}. It should be noted that the first dense layer of the top layer in model 5 is smaller than that of the  other models. This is because model 5 was overfitting if the number of neurons in the first layer was 256, the same number used in the other models. Hence, the size of the dense layer was reduced to reduce overfitting and achieve good performance.

A 2-stage training method was used for the transfer learning model to minimize the error between the ground truth  labels and the predicted labels. This approach is different from the training approach used for Model-A and Model-B which were trained from scratch. In the first stage, the base layer was fixed while the fully connected top layer was trained using the Adam optimizer after being initialized using the glorot-uniform method. 
This was done in order to initialize the weight of the top layer close to the weight of the base layer. Thereafter, the whole model was retrained using the stochastic gradient descent (SGD) with momentum optimizer in order to tune the whole weight of the model appropriately. SGD with momentum was used because it is less aggressive than the adam optimizer as the use of an aggressive optimizer in the second step might cause the information in the base layer to be significantly eroded or lost. The binary categorical entropy was used as the cost function in the entire training process.

\subsection{Experiments}
\label{sec:dataset}
This work seeks to propose a new approach to design and implement deep learning models for distributed systems without compromising on data privacy and security. It achieves this by extracting the most important/critical machine features intelligible yet human unintelligible features from the dataset. These features are then transmitted across the communication network from the edge devices to the edge server where they are aggregated and used to train a classifier. The experimental methods, performance metrics and tools used in validating our proposed framework is explained in this section. 

\subsubsection{\textbf{Experimental method}}
\label{sec:onefixedsnr}
A 2-stage methodology was used to validate our proposed framework and this method is the same irrespective of the type of dataset or model used. In the first stage, the original training set of the original input dataset (uncompressed images) was used to train the classifier. Thereafter, the test set was then used to obtain the needed performance metric in order to set the baseline. 

 In second the stage, the training set of the feature maps (compressed images of the dataset used in stage 1), is used to train the same classifier model. The feature map, which is smaller than the original image by a pre-determined factor, is obtained by passing the original dataset through the encoder of the autoencoder. Thereafter, the performance metric of the classifier is obtained using the test set of the feature maps and the performance compared to the baseline.

\subsubsection{\textbf{Performance Metric}}
\label{sec:fh}
The effectiveness of the framework is assessed using a simple classification task. The test accuracy of the model obtained after the training process is used as the primary performance metric. Furthermore, the effect of our proposed method on the training and testing time, and number of model parameters were also investigated.

\subsubsection{\textbf{Software and Hardware}}
\label{sec:onefixedsnr}
 The design, training and testing of the deep learning models (Autoencoders and CNN Classifiers) were implemented using Keras deep learning framework on TensorFlow backend, running on a NVIDIA Tesla P100-PCIE-16GB GPU.

\section{Results and Analysis}
\label{sec:results}

 The results of the experimental work are presented in this section. The performance of the proposed framework is compared with our baseline using the performance metrics stated in Section~\ref{sec:fh} above. The baseline performance is represented by compression ratio 1 and it is synonymous to using the uncompressed image to test our various models. Furthermore, it should be noted that the vanilla model for the CIFAR10 and ImageNet datasets are different as stated in section ~\ref{sec:dataset} 

\begin{figure}[htbp]
	 \centering
    	 \includegraphics[width=7cm]{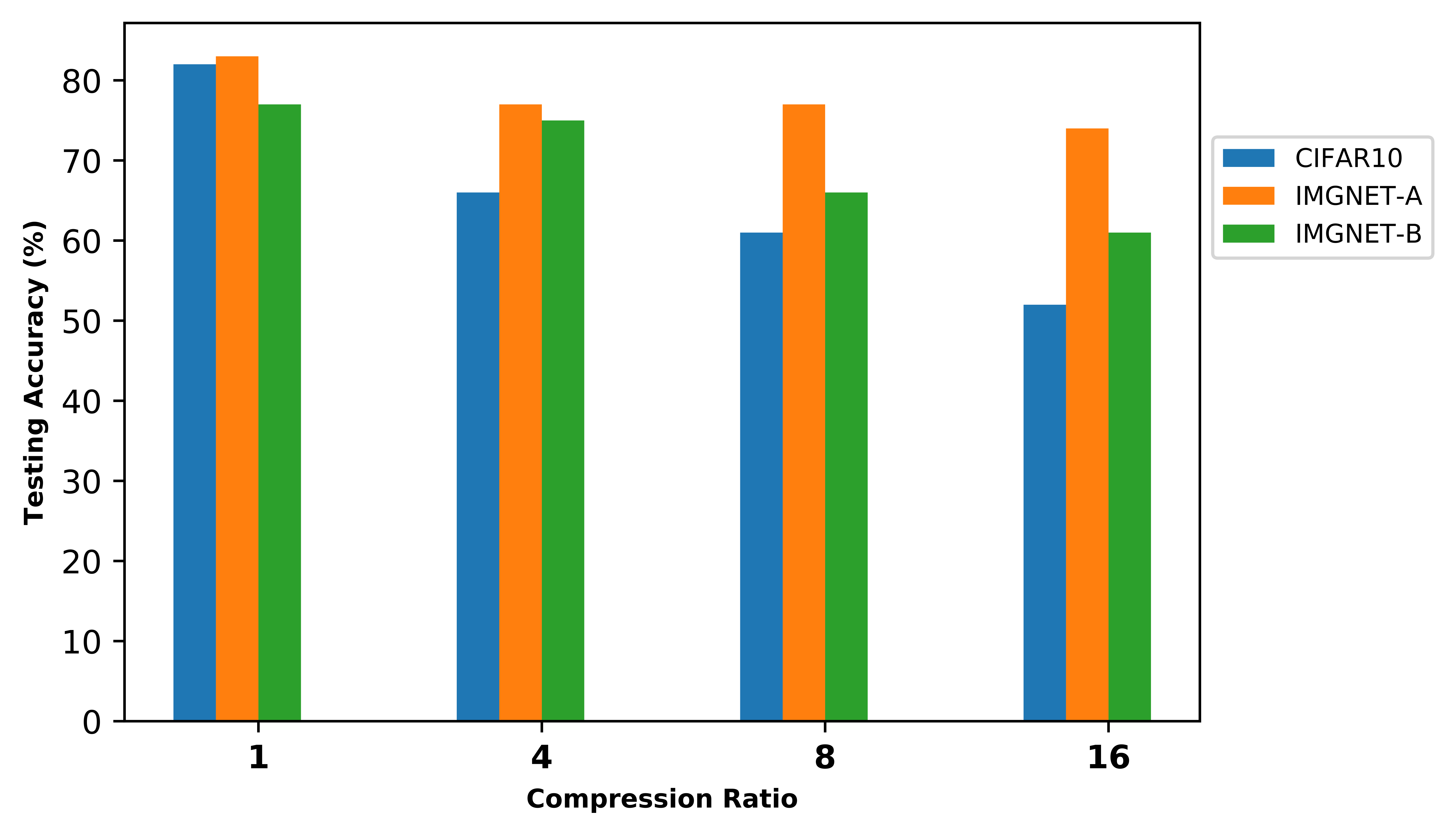}
      	\caption{Comparison of the testing accuracy of the vanilla models for 
	the original dataset (compression ratio =1) and compressed dataset (latent variables) with compression ratio = 4, 8, 16. }
    \label{fig:performanceVanillaModelsCR}
\end{figure}

\subsection{\textbf{Effect on Test Accuracy}}
\label{sec:onefixedsnr}
Figure \ref{fig:performanceVanillaModelsCR} shows the testing accuracy of vanilla CNN Classifiers (Model-A and Model-B) when trained and tested with compressed and uncompressed CIFAR10 and Imagenet datasets.
The testing accuracy for the compression ratio 1 (uncompressed images), representing our baseline, is highest across all the cases, as expected. This is because all features in the raw images can be used for classification.
Furthermore, the testing accuracy for IMGNET-A is higher than that of IMGNET-B. The differences in performance can be attributed to the very close similarity in the images in IMGNET-B, as classifying such images is a much more difficult classification task as compared to classifying images in IMGNET-A. 

A general degradation in the testing accuracy is observed in Figure \ref{fig:performanceVanillaModelsCR}  as the compression ratio is increased, although the rate of reduction varies across the models used for the 3 datasets. The rate of degradation of the testing accuracy of the model tested for CIFAR10 dataset is the highest for all the compression ratios. This is because of the small dimension of the CIFAR10 images (32*32), implies that the amount of features needed to perform a classification task is even smaller when compressed. 

Furthermore, the rate of degradation of the testing accuracy for IMGNET-A dataset was very modest across all the compression ratios. However, similar performance was not observed in IMGNET-B, particularly for compression ratios 8 and 16 despite having the same image dimension (256*256). The bigger degree of degradation observed in the case of IMGNET-B for compression ratios 8 and 16 is due to the complexity of the classification task. This is because of the similarities in the images that make up the various classes in IMGNET-B, unlike  IMGNET-A where the the images that make up the classes are very different.Hence, the complexity of its classification task  means it requires a lot more features than that of IMGNET-A. Furthermore, bigger compression ratio also means the model has less amount of features to make a classification decision. 

\begin{figure}[htbp]
	 \centering
    	 \includegraphics[width=7cm]{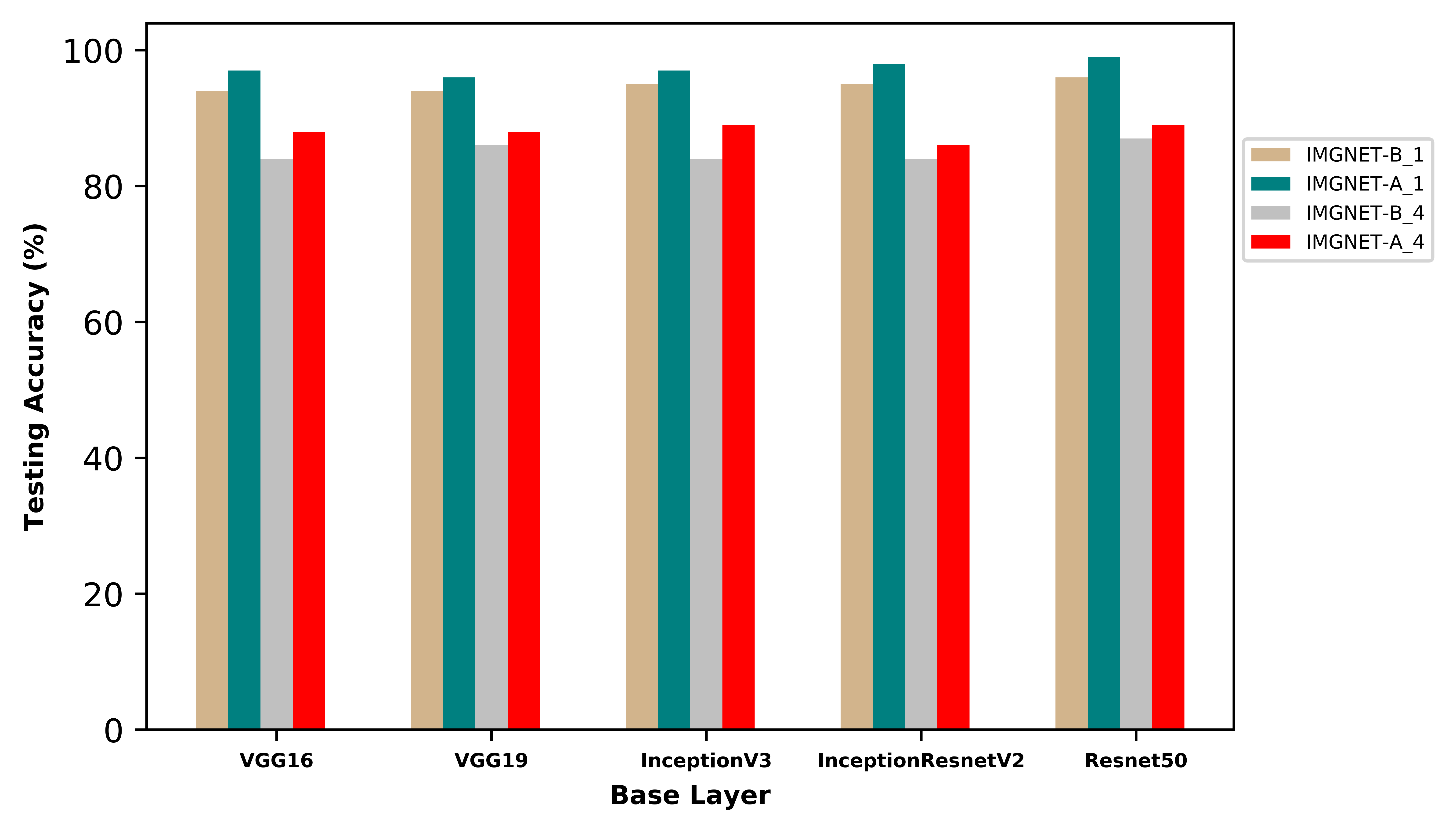}
      	\caption{Testing accuracy of the transfer learning based model (Model-C) using different base models for the ImageNet dataset with compression ratio = 4. }
    \label{fig:performanceModelC_CR}
\end{figure}

The testing accuracy of transfer learning based model (Model-C) for the ImageNet dataset compressed by a factor of 4, using different base models, is shown in Figure \ref{fig:performanceModelC_CR}. The transfer learning model was not designed and trained using the CIFAR10 datasets as its performance was poor with the compressed images. The poor performance can be attributed to the very deep nature of the transfer learning based model leaving inadequate number of features available at the beginning of the fully connected layer (top layer) where classification takes place. The same reason also explains why the transfer learning model was only designed and tested with ImageNet dataset with compression ratio 1 and 4 only.

The testing accuracy of the transfer learning model across different base models for IMGNET-A and IMGNET-B datasets at compression ratio 1 (baseline) and 4 is higher than the corresponding performance of the vanilla model (Model-A and Model-B). This performance can be attributed to the powerful feature extraction property of the different base layer used. 
However, the rate of degradation in the testing accuracy for compression ratio 4 is  higher than what was observed for the vanilla models. This can also be attributed to the very deep nature of the transfer learning models as a small amount of information/features left is not distinct enough to make accurate classification.

\subsection{\textbf{Effect on Number of Parameters}}
\label{sec:onefixedsnr}

\begin{figure}[htbp]
	 \centering
    	 \includegraphics[width=7cm]{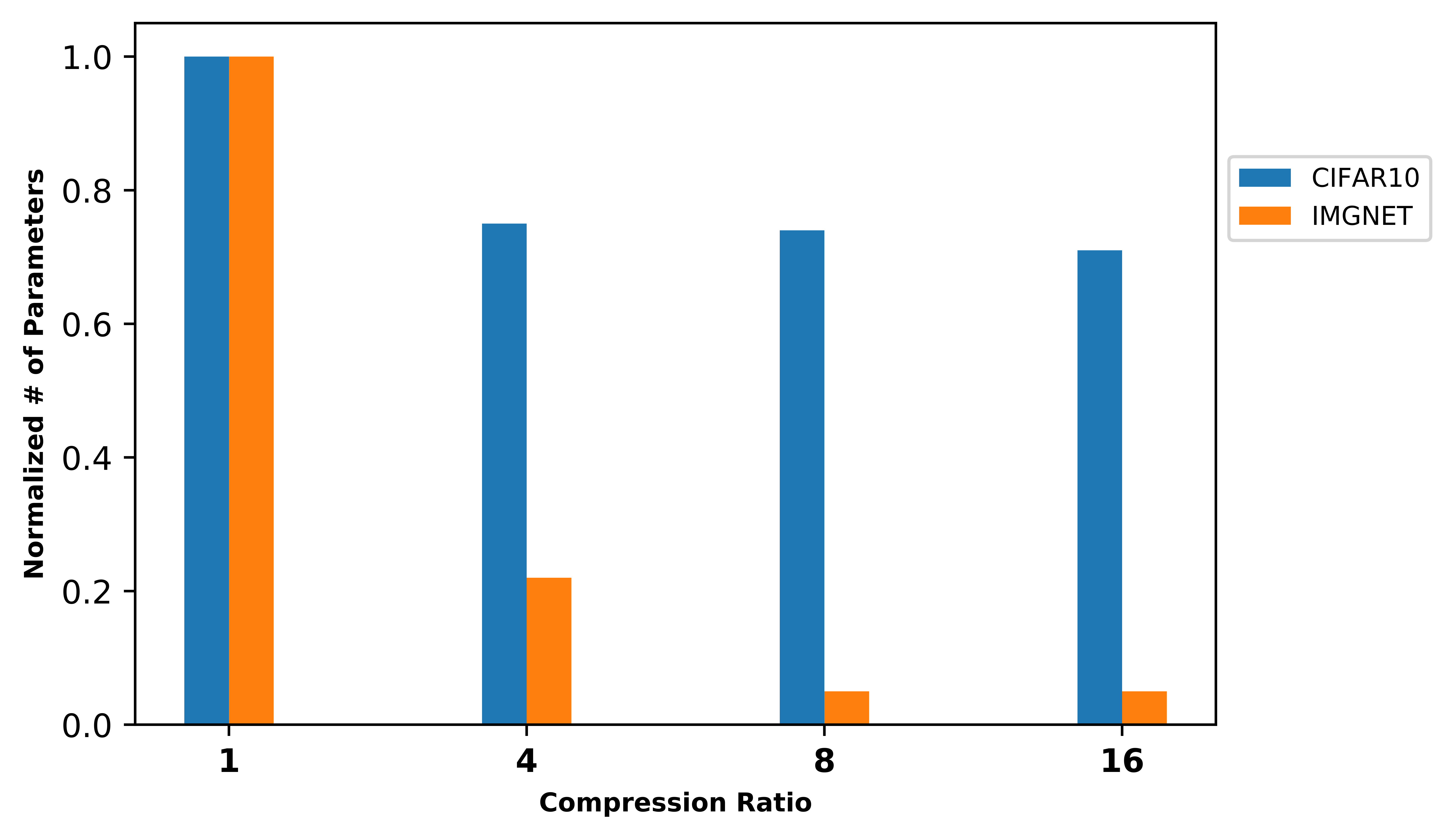}
      	\caption{Comparison of the normalized number of vanilla model parameters vs. data compression ratio}
    \label{fig:performanceNumberParameters}
\end{figure}

The number of parameters in a convolutional neural network is determined by many factors such as the filter size, the number of filters, the size of the input data, the number and type of hidden layers, etc. Hence, reduction in the number of parameters can be achieved by reducing the size of the input data. 
The relationship between the normalized number of parameters in the vanilla model for the CIFAR10 and ImageNet datasets vs. the compression ratio is shown in Figure \ref{fig:performanceNumberParameters}.
It can be observed that for the same compression ratio, the rate of reduction in the normalized number of parameters of the vanilla model for the ImageNet dataset is bigger  than that for the CIFAR10 dataset. The bigger rate of reduction can be attributed to the size of the image, and in turn, the number of features, which impacts the number of parameters in the fully connected layers of the model. 
It was also observed that the rate of reduction in the normalized number of parameters for each model appears to be flat after a certain compression ratio which varies from model to model. The flatness is due to the reduction in the significance of the fully connected layer to impact on the number of parameters, as the number of features reduces below a certain point.

\subsection{\textbf{Effect on Testing and Training Time}}
\label{sec:onefixedsnr}

\begin{figure}[htbp] 
	\centering
		\includegraphics[width=7cm]{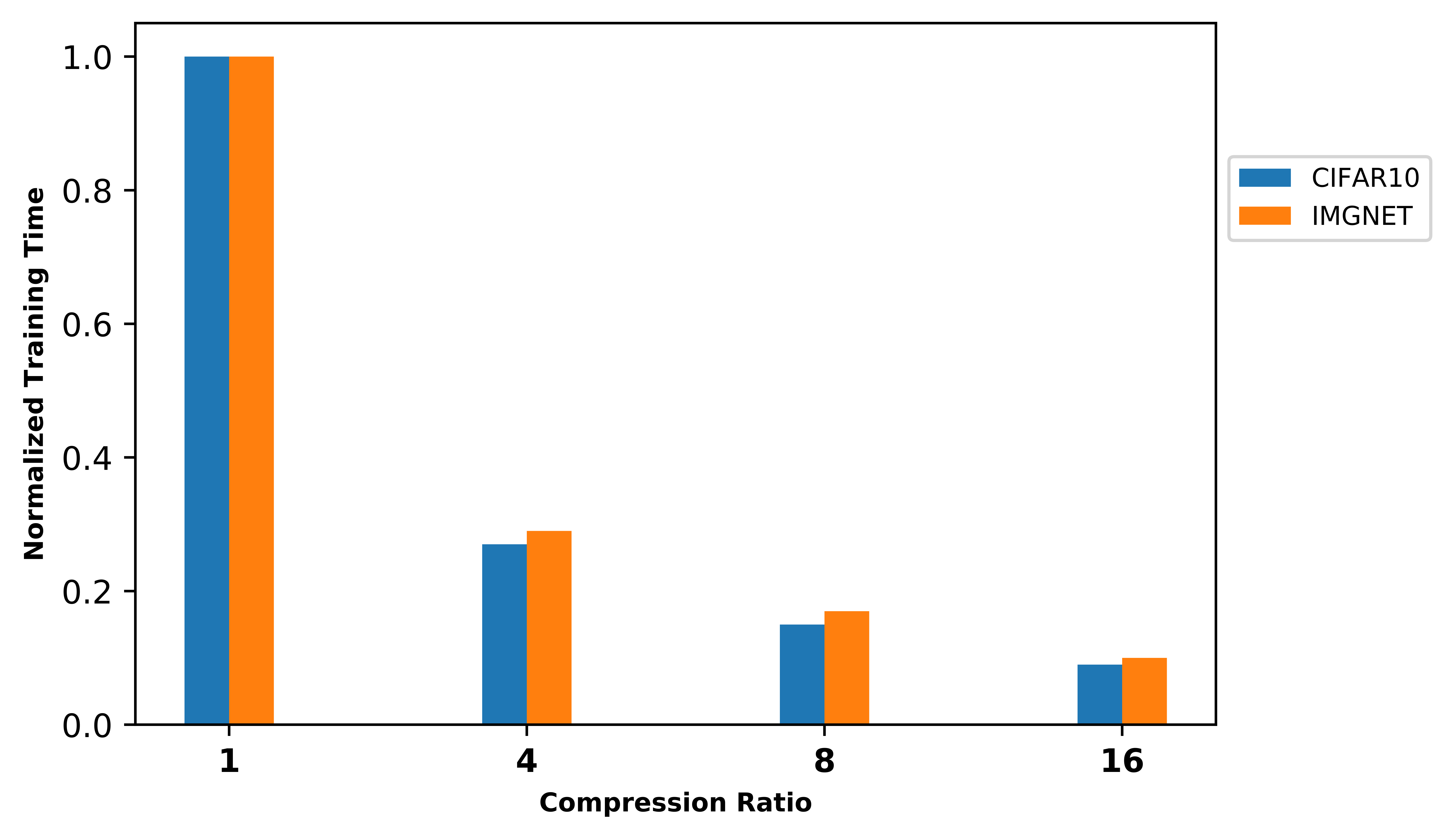}  \\
		(a) Comparison of the normalized testing time  \\
	\vspace{0.05in}
		\includegraphics[width=7cm]{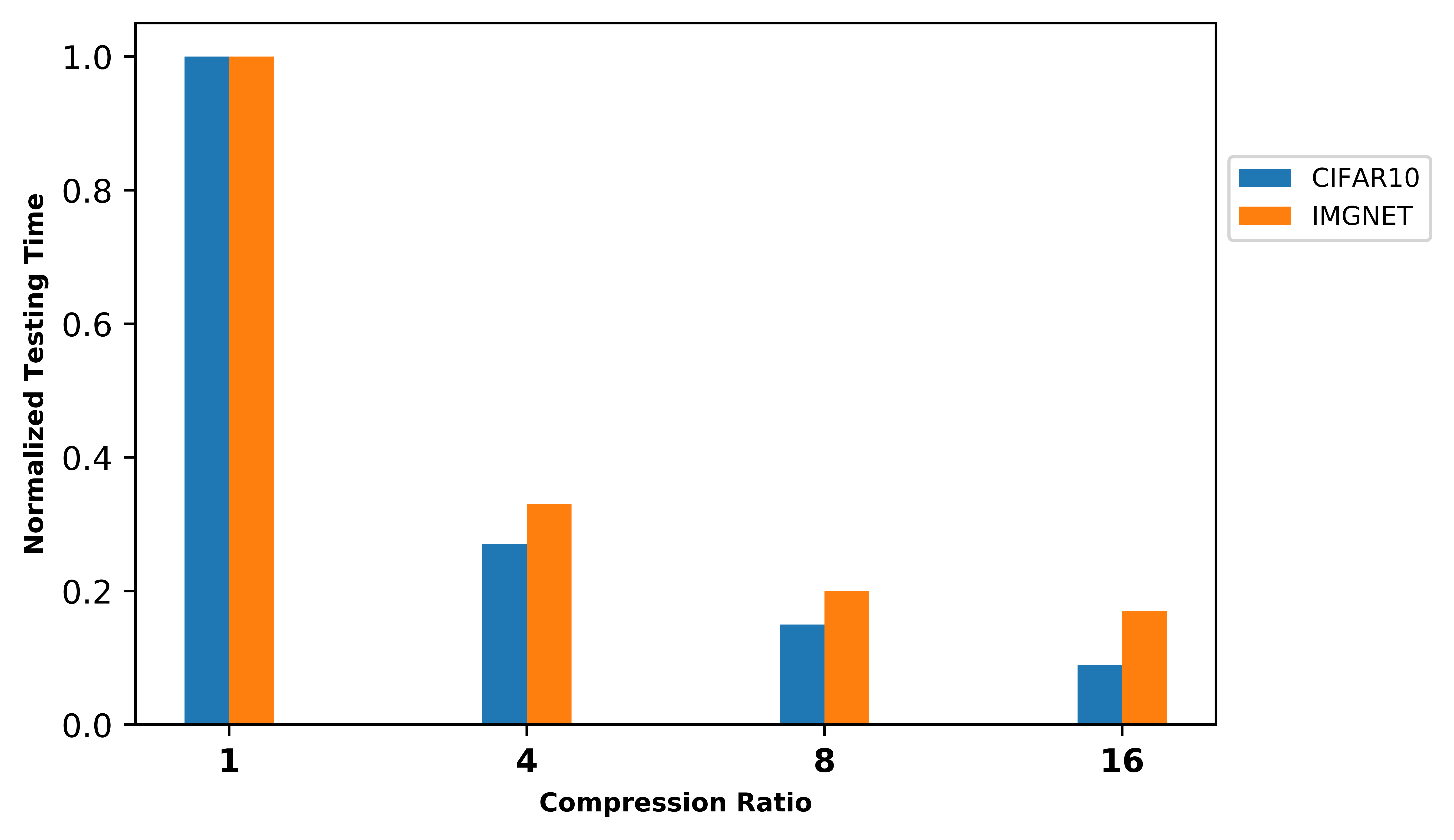}  \\
		(b) Comparison of the normalized training time  \\
	\caption{Comparison of the normalized testing time and training time of the vanilla models for various compression ratios for CIFAR10 and ImageNet datasets.   }
	\label{fig:performanceTestTimeTrainTime}
\end{figure}

Figure~\ref{fig:performanceTestTimeTrainTime} show the normalized amount of time required for testing and training, respectively, of the vanilla models for various compression ratios for CIFAR10 and ImageNet datasets. A reduction in the amount of training and testing time was observed across the compression ratios and the models. The reduction can be attributed to the decreasing size of the input data and smaller number of parameters to learn, when compression ratio increases.

\section{Discussion and Related Works}
\label{sec:discussion}
There are several methods proposed in the literature to address the privacy and security concerns associated with data used for training deep learning models. Examples of popular approaches include homomorphic encryption\cite{Xie2014CryptoNetsNN},  differential privacy \cite{Dwork1,Abadi2016DeepLW} and secure multiparty computation \cite{Liu2017ObliviousNN}. Despite the successes of these methods, some issues remain such as performance degradation, non-trivial overhead or limited application~\cite{Bae2018SecurityAP,ZhaoDifferentialReview,Shokri}.

The use of collaborative deep learning method, such as federated learning, has been introduced in recent years to solve the problem of data privacy. Federated learning is a type of machine learning where the goal is to train a high quality centralized model while the data remains distributed over a large number of clients~\cite{McMahanMRA16}. It involves the sharing of model parameters and model gradients through a parameter server without sharing their local data. Federated learning is based on an  iterative model averaging and it is robust to unbalanced data and non-i.i.d. data distribution. Federated learning has been applied to mobile keyboard prediction, vocabulary word learning and google keyboard query suggestions improvement~\cite{YangTimothy2018,HardAndrew2018,ChenMin2019}.
Federated learning may be viewed as an extension of the idea discussed in  ~\cite{dean,ChenJ} that stochastic gradient descent can be implemented in parallel and asynchronously.

Federated learning may suffer from non-trivial communication cost. To deal with the high communications cost, an efficient multi-objective evolutionary algorithm, based on a scalable network connectivity encoding method, was proposed in~\cite{MultiObjFederatedLearning}.
To help reduce the uplink communication bottleneck, the use of structured and sketched updates were introduced in~\cite{KonecnyFederated}.

Federated learning may also suffer from security/privacy issues due to the need to communicate the model parameters to the central server. One recent study showed potential security/privacy issues due to the possibility of reconstructing original data  from the shared gradient~\cite{leakingGra}. 
Secure aggregation, a type of secure multi-party computation algorithm for federated learning was introduced in~\cite{Bonawitz2017PracticalSA}. This helps guarantee the privacy of data used in generating gradients shared by each model and improve communication efficiency. Furthermore, it was observed that federated learning performs poorly when the data distributed across the training center is strictly non-i.i.d. of a single class. This statistical challenge was resolved by creating and using a small subset of data which is globally shared between all the edge devices~\cite{ZhaoFLwithNonIID2018} or adopting a multi-task learning approach~\cite{Smith2017} .

Autoencoder has been applied to address data privacy concerns in several recent works~\cite{Mirjalili_2018,ALGULIYEV20191,AutoencoderNeurology2018}. In~\cite{Mirjalili_2018}, a convolutional autoencoder that perturbs an input face image to impart privacy
to a subject is proposed. It is shown the method can protect gender privacy of face images.
A proof-of-concept study has been performed in~\cite{AutoencoderNeurology2018} to use an autoencoder for preserving video privacy, especially when non-healthcare professionals are involved. A modified sparse denoising autoencoder has been applied in~\cite{ALGULIYEV20191} to reduce the sparsity and denoise the data. Then a 3-class classification is performed on the reconstructed data from the autoencoder and it is shown that the classifier can classify the original black class data as the transformed gray class data. 

Although autoencoder has been used to address data privacy concerns, this work is the first in the use of autoencoder for addressing privacy concerns, communication cost, and deep learning efficiency associated with mobile edge computing systems with large number of edge devices.
This is achieved by using the autoencoder to extract human unintelligible but machine intelligible features from the data. The features or latent vectors are then used to train the classifier. Furthermore, the proposed approach comes with the added advantage of reducing the dimensionality of data needed to be transmitted, thus reducing the communication cost and the number of model parameters, as well as training and inference time. 
This approach does not suffer from leaking gradient problem associated with federated learning~\cite{leakingGra}.

\section{Conclusions}
\label{sec:conclusion}
A novel edge computing framework for designing and implementation of privacy preserving image classification models is proposed in this work. The proposed framework provides 1) flexibility of training autoencoder at each edge device individually, thus protect data privacy of end users because raw data is not transmitted at any time; 2) after the training of autoencoder is complete, raw data is ``compressed'' and ``encrypted'' by the encoder before transmitting to the edge server, and this will reduce the communications cost and further protect the data privacy and security; 3) the autoencoder will provide features to the classifier at the server, thus allow the classifier to be trained on the features with less and controlled dimensions; 4) the decoupling of the training of the autoencoder at the edge devices and the training of the classifier at the edge server relaxes the frequent communications requirement between edge devices and edge server. 
Experiments have been carried out using CIFAR10 and ImageNet datasets, and detailed analysis of tradeoff between classifier accuracy, dimensionality of data, compression ratio and different choice of classifiers has been given to provide benchmark and insights on the proposed scheme.

For future work, comparison with federated learning in terms of classifier performance vs the communications cost and model complexity will be carried out for image classification tasks. This will help quantify the pros and cons of the proposed approach.
Furthermore, the use of other types of autoencoder to extract latent variables and use of knowledge distillation to help mitigate the reduction in the model accuracy will be explored.

\section{Acknowledgments}
\label{acknowledgement}
This research work is supported in part by the U.S. Dept. of Navy under agreement number N00014-17-1-3062 and the U.S. Office of the Under Secretary of Defense for Research and Engineering (OUSD(R\&E)) under agreement number FA8750-15-2-0119. The U.S. Government is authorized to reproduce and distribute reprints for governmental purposes notwithstanding any copyright notation thereon. The views and conclusions contained herein are those of the authors and should not be interpreted as necessarily representing the official policies or endorsements, either expressed or implied, of the U.S. Dept. of Navy  or the U.S. Office of the Under Secretary of Defense for Research and Engineering (OUSD(R\&E)) or the U.S. Government.

\bibliographystyle{IEEEtran}
\bibliography{AutoencoderCNN}

\end{document}